\documentclass[letterpaper, 10 pt, conference]{ieeeconf}  

\IEEEoverridecommandlockouts                              

\usepackage{graphicx}
\usepackage{amsmath}
\usepackage{amsfonts}
\usepackage{amssymb}
\usepackage{gensymb}
\usepackage{booktabs}
\RequirePackage[colorlinks=true, allcolors=blue]{hyperref}
\usepackage{orcidlink}  

\usepackage[
    backend=biber,
    style=ieee,
    natbib=true,
    doi=false,
    url=false,
    isbn=false,
]{biblatex}

\AtEveryBibitem{%
   \clearfield{Title}%
   \clearfield{eprint}%
   \clearfield{note}%
}

\renewcommand{\citet}[1]{\textcite{#1}}
\renewcommand{\citep}[1]{\parencite{#1}}

\newcommand{\printglossaries}{}

\title{Advancing Monocular Video-Based Gait Analysis Using Motion Imitation with Physics-Based Simulation}

\author{
Nikolaos Smyrnakis$^{{1}}$ \and Tasos Karakostas$^{{1, 2}}$ \and R. James Cotton \orcidlink{0000-0001-5714-1400} $^{{1, 2}}$ 
\thanks{
This work was supported by the Research Accelerator Program of the Shirley Ryan AbilityLab and the Restore Center P2C (NIH P2CHD101913).
}
\thanks{
$^{{1}}$ {Shirley Ryan AbilityLab}
} 
\thanks{
$^{{2}}$ {Northwestern University}
} 
}

\begin{document}

\maketitle


\begin{abstract}
Gait analysis from videos obtained from a smartphone would open up many clinical opportunities for detecting and quantifying gait impairments. However, existing approaches for estimating gait parameters from videos can produce physically implausible results. To overcome this, we train a policy using reinforcement learning to control a physics simulation of human movement to replicate the movement seen in video. This forces the inferred movements to be physically plausible, while improving the accuracy of the inferred step length and walking velocity.\\
\end{abstract}


\section{Introduction}

Easier access to gait analysis would be a powerful tool for rehabilitation care and research. Ultimately, this could enable monitoring gait in all clinic settings and the home, allowing early identification and intervention for gait impairments, and reducing the incidence of adverse outcomes like falls. While human pose estimation (HPE) algorithms are advancing rapidly, less work has focused on estimating clinically relevant features such as spatiotemporal gait parameters. We have shown that a transformer trained on a large dataset from a clinical gait laboratory can accurately estimate gait event timing and also step length and velocity  \citep{cotton_transforming_2022}, which we then validated prospectively in the clinic \citep{cimorelli_portable_2022}. However, directly regressing kinematics from keypoints does not enforce physical or biomechanical constraints during inference. This approach sometimes results in inconsistencies, such as the velocity not being equal to the average step length times the cadence.

A potential solution to these implausible results is integrating physics-based simulation into the inference process, with forces applied to the simulated model by a policy learned through reinforcement learning. Early examples of this work focused on training a policy for single movement measured with optical motion capture \citep{peng_deepmimic:_2018} and video \citep{peng_sfv:_2018} and were even trained for muscle-actuated biomechanical models\citep{lee_scalable_2019}. The model mismatch between human movements and simulated bodies makes training policies for general motion imitation challenging. This barrier can be mitigated using residual force control with an additional non-physical force applied to the model pelvis \citep{yuan_residual_2020}. By accounting for this, \citet{yuan_simpoe_2021} (SimPoE) learned a policy that tracks movements from video. The impressive results showed that the physics simulation reduced artifacts such as jerky, discontinuous movements, feet penetrating through the floor, or sliding on the ground. Our method is heavily inspired by SimPoE. Subsequent work has shown that physics-based imitation can replicate all the motions captured in a large dataset of human movements \citep{luo_universal_2023, luo_perpetual_2023}. Like most HPE algorithms, these works focus on the 3D joint location accuracy. These algorithms also assume the anthropomorphic measurements are given a priori. Less attention has been given to joint angles, testing on diverse, clinical populations, or the accuracy of spatiotemporal gait parameters.

Using reinforcement learning to train policies to produce forces that imitate a diverse set of observed movements requires millions or billions of simulation steps. This has become feasible in recent years with the development of several simulators that can run many accelerated and parallel simulations on the GPU including Brax \cite{freeman_brax_2021} and IsssacGym \citep{makoviychuk_isaac_2021}. A very mature simulator that supports some biomechanical models is Mujoco \citep{todorov_mujoco_2012}, which very recently provided GPU acceleration based on a new physics simulator that emerged from Brax. Excitingly, multiple biomechanically oriented environments building on Mujoco have recently been released including MyoSuite \citep{caggiano_myosuite_2022} and a locomotion imitation benchmark suite \citep{al-hafez_locomujoco_2023}. Our work started before these recent developments, so we use a simplified humanoid model in Brax. We anticipate that our findings will naturally translate to and benefit from these newer GPU-accelerated biomechanical models.

The goal of this work is to explore incorporating physics-based simulation into the inference process for gait analysis. This involves handling a wide diversity of gait patterns and body sizes. We find this approach promising and that it improves the accuracy of our estimates of walking velocity and step length compared to our prior approach.

\section{Methods}

\subsection{Gait Laboratory Dataset}

This study was approved by the Northwestern University Institutional Review Board. The dataset we used for training and validating this algorithm was previously described in \citep{cotton_transforming_2022}. This dataset jointly recorded video and marker-based motion analysis from subjects with a wide range of ages, heights, and range of gait impairments, with a heavy emphasis on children with cerebral palsy.

\subsubsection{Video processing}

Videos recorded at 30 fps were processed using PosePipe \citep{cotton_posepipe_2022}, using the same annotations to identify the person undergoing gait analysis and approach to synchronized motion capture (mocap) data to the videos we previously described \citep{cotton_transforming_2022}. All videos were obtained in the frontal plane. Keypoints in 3D were estimated independently for each frame using an algorithm trained on numerous 3D datasets that outputs the superset of all these formats (MeTRAbs-ACAE) \citep{sarandi_learning_2023}. We used a subset of the 87 keypoints from the MOVI dataset \citep{ghorbani_movi_2021}, which correspond to those often used in biomechanical analysis. These are the same keypoints that we use in our markerless motion capture pipeline \citep{cotton_markerless_2023, cotton_improved_2023}, that we have found to provide high-quality estimates.

To limit the input dimension, particularly when including multiple future observations, we used 23 keypoints from the full 87. Specifically, we included bilateral keypoints for the heels, big toes, feet, ankles, knees, hips, elbows, shoulders, wrists, and hands as well as the head, pelvis, and sternum. The skeleton trajectories were normalized by aligning the average vector from the right to the left hip with the x-axis and the average vector from mid-hip to the thorax keypoint with the y-axis. MeTRAbs-ACAE estimates keypoints in absolute world coordinates, so we also centered the keypoints at the mid-hip location. Keypoints were rescaled from mm to meters for consistency with the Brax units. Thus each trial, $i$, had a keypoint trajectory $\mathbf k_i \in \mathbb R ^{T \times 23 \times 3}$, where $T$ is the duration of the sequence.

\subsubsection{Marker-based gait analysis}

Gait analysis had previously been performed following a standard clinical workflow. From this, we extracted the hip, knee, and ankle joint angles. We also extracted the hip, knee and ankle joint kinematics, specifically the 3D joint velocities relative to the forward orientation of the participant and the 3D joint locations relative to the mid-pelvis. Thus each trial had a mocap trajectory, $\mathbf m_i \in \mathbb R^{T \times 18 \times 3}$.

\subsubsection{Trial selection}

We restricted our dataset to trials that included at least 75 frames of videos that overlapped with the motion capture data and where the subject was identified by MeTRAbs-ACAE in at least 95\% of the frames. We also excluded any trials any joint angles that exceeded 136$\degree$, or where the height was not extracted from the original database. We had previously split subjects into training and testing subjects \citep{cotton_transforming_2022}. This resulted in 141 subjects with 1099 trials in our test split. We kept the first 4200 trials (from 481 subjects) for training and used the remaining 822 trials (from 99 subjects) as validation data during hyperparameters tuning.

\subsection{Brax model design}

Our humanoid model was based on the one available in Brax \citep{freeman_brax_2021}, which in turn is based on \citet{tassa_synthesis_2012}. We made several modifications to this model. First, we removed the upper half of the torso, as our gait data does not include tracking of the arms or torso. Then, we modified the legs to include feet with a single-axis joint at each ankle for plantarflexion and dorsiflexion. We included residual forces at the torso with 3 degrees of freedom for lateral forces and 3 degrees of rotational forces, following \cite{yuan_residual_2020}. We also normalized the pelvis width, thigh length, and leg length per meter of height to the median value found in our dataset. During training and inference, the joint positions and velocities were scaled by the height of the participant, and these scaled values were used when computing the reinforcement learning reward function, defined further below.

The state of the model includes the current body segment angles, body segment rotation velocities, joint positions, and joint velocities, $\mathbf s=(\mathbf q, \mathbf {\dot q, \mathbf x, \mathbf {\dot x}})$. In the Brax V1 physics engine, the body segment angles and rotations are represented in the world reference frame. The model includes an inverse kinematic function to recover joint angles and velocities, $\phi=FK_\phi^{ -1}(\mathbf q, \mathbf {\dot q})$ and $\dot \phi=FK_{\dot \phi}^{ -1}(\mathbf q, \mathbf {\dot q})$, which we use in the reward function when comparing to the mocap joint angles.

The model provides a physics-based transition function to predict the next state given a current state and action: $\mathbf s_{t+1} = \mathcal T (\mathbf s_t, \mathbf a_t)$. The action space is the set of joint torques and the 6-degree-of-freedom residual force applied to the pelvis.

The states, actions, and rewards were computed at 30 Hz (the video framerate, and what the ground truth data was synchronized to). The physics engine took 16 substeps per video frame.

To account for potential model differences between the simple humanoid model and the clinical biomechanical model, we also included a learned offset for each sagittal plane joint angle but found this was typically less than 1 degree of offset.

\subsection{Trajectory following environment}

We modified the environment to include the set of training trajectories which included the keypoints, $\mathbf k_i$, along with the mocap data, $\mathbf m_i$, which was only used in computing the rewards. At each environment reset, a random trajectory was selected for the policy to replicate.  We posed the trajectory tracking problem as a goal-conditioned reinforcement learning problem, where a policy produces an action conditioned on the current state and observations that provide the goal $~\mathbf a \sim \pi(\mathbf s, \mathbf o)$. The observations passed to the policy could include multiple future timesteps, $F$, of information concatenated together to allow the policy to anticipate future tracking: $\mathbf o_i^{(t)} = [\mathbf k_i^{(t)}, \mathbf k_i^{(t+1)}, \ldots, \mathbf k_i^{(t+F)}]$. Environment rollouts were terminated when there were no future observations, if the reward became negative, or if the model fell over.

To compute the reward, we first estimated the errors between the ground truth measurements and the model state. This was divided into errors for the joint angles focusing on the sagittal plane, $e_\phi \in \mathbb R^6$, joint positions, $e_x \in \mathbb R^{6 \times 3}$, and joint velocities, $e_{\dot x}\in \mathbb R^{6 \times 3}$. We focused on only sagittal plane joint angles as we anticipated the vast differences between our simplified humanoid model and the accurate biomechanical model used in the mocap data would be too significant of a model mismatch. Joint positions were computed related to the pelvis, instead of being evaluated in the world reference frame, to avoid overpenalizing errors from earlier in a trial. Velocities remained in the world reference frame. From this, we computed two reward functions. First, following SimPoE \citep{yuan_simpoe_2021}  and other physics-based tracking papers, we used the product of exponential errors for each of the modalities:

\begin{equation}
r_{\mathtt {exp}}(\mathbf s_t, \mathbf a_t) = \alpha_0 \cdot e^{-\alpha_a \| e_a \|^2} \cdot e ^ {-\alpha_x \sum_i \| e_x^{(i)} \| ^2} \cdot e ^ {-\alpha_{\dot x} \sum_i \| e_{\dot x}^{(i)} \| ^2}
\end{equation}

Where $e^{(i)} \in \mathbb R^3$ are the errors for each joint, and $\alpha_a=5$, $\alpha_x=2.5$ and $\alpha_{\dot x}=0.17$ are the weights for the joint angles, joint positions, and joint velocities, respectively.  The $\sum_i \| e_x^{(i)} \| ^2$ operations indicate taking the 3D Euclidean norm for each joint position and velocity error before summing them together.

This type of exponential reward is typical in the motion imitation literature \citep{yuan_simpoe_2021}. We compared this to a reward function based on the mean squared error between the ground truth and model state:

\begin{align}
r_{\mathtt {mse}}(\mathbf s_t, \mathbf a_t) &= \alpha_0 -\alpha_a \| e_a\|^2 - \alpha_x \sum_i \| e_x^{(i)} \| ^2 \\
 &\qquad - \alpha_{\dot x} \sum_i \| e_{\dot x}^{(i)} \| ^2 - \beta \| a_t \|^2
\end{align}

. We used $\alpha_0=8$, $\alpha_a=2.5$, $\alpha_x=3.9$, $\alpha_{\dot x}=20$, and $\beta=0.005$ for all experiments.

\subsection{Reinforcement learning and control policy}

We used a standard reinforcement learning approach to train an observation-conditioned action policy, $\mathbf a \sim \pi_\theta(\mathbf s, \mathbf o)$, which is trained to maximize the expected reward with a discount factor $\gamma$:

\begin{equation}
\mathcal J(\pi_\theta) = \mathbb E_{\mathbf o, \mathbf y \sim \mathcal D,\mathbf a, \mathbf s \sim \pi_\theta(\mathbf s, \mathbf o)} \left[  \sum_{t=0}^T \gamma^t r(\mathbf s_t, \mathbf y_t) \right].
\end{equation}

We trained our model with proximal policy optimization (PPO) \citep{schulman_proximal_2017}, using the implementation provided with Brax. This was modified to optionally allow propagating gradients through timesteps during rollouts to allow training an initializer network, described further below. We also implemented short horizon actor-critic in Brax as an optimizer, given the claim it can produce more precise tracking for differentiable environments \citep{xu_accelerated_2022}. However, we found that it was extremely slow to train and could not obtain good policies.

In our standard formulation, the actions from the policy were directly applied as torques or residual forces to the humanoid model. Physics-based imitation learning typically uses a proportional derivative (PD) controller \citep{peng_deepmimic:_2018, yuan_residual_2020, yuan_simpoe_2021, luo_universal_2023, luo_perpetual_2023}, where the setpoint and gains are learned through reinforcement learning. We compared direct control, as just described, with PD-based control. In PD control, the policy outputs:

\begin{equation}
\mathbf u, \mathbf k_p, \mathbf k_d \sim \pi^{\mathtt PD}_\theta(\mathbf s, \mathbf o)
\end{equation}

and then the action is computed as:

\begin{equation}
\mathbf a = \mathbf k_p \cdot \left (\mathbf u - FK^{-1}(\mathbf q, \mathbf {\dot q}) \right) - \mathbf k_d \cdot FK_{\dot \phi}^{-1}(\mathbf q, \mathbf {\dot q})
\end{equation}

In this case, we directly obtained the 6 degrees of freedom for the residual force from the policy and applied them to the humanoid.

\subsection{Augmentation}

During training, we applied data augmentation by adding random scaled Gaussian noise to the keypoint observations. We also used random noise to initialize the joint angles after resetting the environment.

\subsection{Initializer}

Typically, at the beginning of a rollout the initial value of $\mathbf s$ is reset independently of the trajectory, possibly with the addition of random noise. However, this produces large errors at the beginning of tracking. To reduce this, we implemented an initializer network that would output the desired initial pose on environment reset based on the initial keypoints, $\mathbf k_i^{(0)}$. This required modifying the PPO algorithm to pass the gradients through the timesteps, which approximately doubled the training time, so this was not used in the majority of our experiments.

\subsection{Hyperparameter Tuning}

There were a large number of hyperparameters we explored. These included the optimizer (Adam \citep{kingma_adam_2015}\} versus RMSProp), the learning rate, the reward (exp or MSE), and $\alpha_0$, using a PD controller versus direct control, learning rate scheduling, initialization noise, the number of future observations to provide, policy and value network architectures, the PPO epsilon clipping, and the amount of augmentation applied. This was partially performed through Bayes optimization through Weights and Biases \citep{wandb}, and then was refined through manual parameter adjustments and inspecting the results. A large number of parameters produced very poor results or took a very long time to learn. After optimizing the hyperparameters for both the ground truth and keypoint-driven models, we then trained the model on the full training dataset, including the trials used for validation, and computed the final metrics on the test data.

To see the impact of several implementation decisions, we include the performance of specific model variations during the development when also applied to the test data set. These models were not trained with the validation data (thus are expected to slightly underperform our final model) and these results were not used to influence our parameter selection.

Models were trained using 4096 parallel environments on a mixture of A6000s and A100s. Most models were trained for 1 billion total simulation steps, which would take approximately 4 hours.

\subsection{Final implementation details}

Our final model used a policy depth of 8 layers and a width of 256 layers. The value function had a depth of 8 layers and a width of 1024 layers. The PPO entropy cost was 0.003 and epsilon clipping was 0.2. We used 4 future steps of observations. The reset noise scale was 0.15 and the observation augmentation noise scale was 0.0001. We used a reward discount of 0.99. We simulated 4096 parallel environments, and policy updates used a batch size of 1024 with 128 minibatches.

\subsection{Performance Metrics}

Because of the number of frames for the model to start tracking well, we only scored the trajectories after the first 45 frames. To quantify the performance tracking the hip and knee, we measured the root mean squared error (RMSE) for each of the joint angles in degrees.

In addition to reporting the RMSE of each of the joint angles, positions, and velocities, we also computed several gait metrics. These included the walking velocity, step length, and step width. Walking velocity was computed over a single gait cycle determined by the ground truth mocap gait events and step length and step width were extracted at the time of each foot contact. For each metric, we computed the correlation coefficient and the root mean squared error. We also show Bland-Altman plots \citep{bland_statistical_1986} and reported the interquartile range.

\section{Results}

First, we show the results from our best-performing model. Model parameters were selected from performance on the validation data and then a model was trained on the entire training dataset. When training this model, we also included the initializer network which sets the initial pose based on the keypoints, with all body segments having zero velocity.

\subsection{Example waveforms}

\begin{figure}[!htbp]
\centering
\includegraphics[width=1\linewidth]{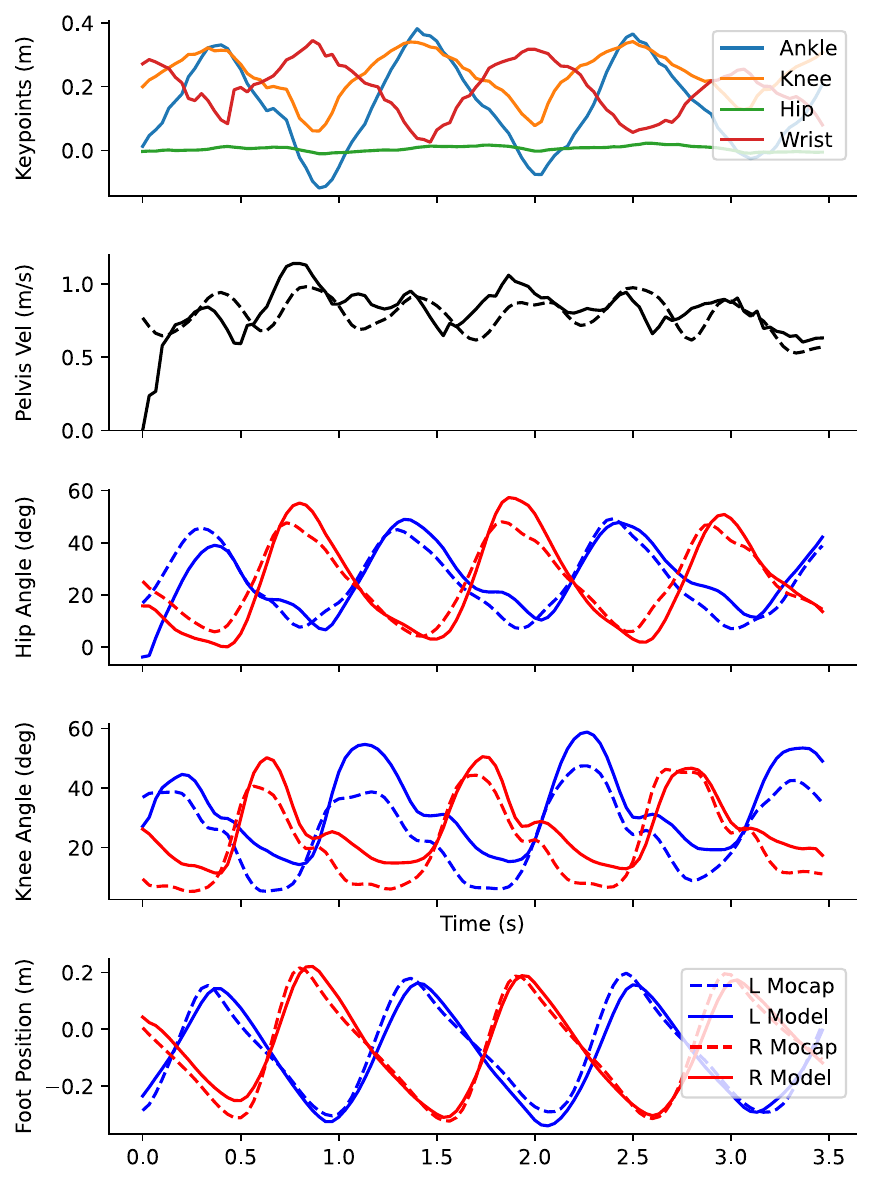}
\caption[]{Waveforms from a single trial. The top row shows the sagittal plane position of several keypoints on the left side. The remaining rows show the pelvis velocity, hip angles, knee angles, and ankle positions. The mocap data is shown with dashed lines and solid lines are the model predictions.}
\label{fig:rollout}
\end{figure}

Figure~\ref{fig:rollout} shows waveforms from a sample trial. This example shows a fairly good, but not perfect, performance reproducing the motion capture data using the policy driven by the keypoints estimated from video frames. It also demonstrates some lag behind the target waveforms, which was a common occurrence. Because the initializer only sets the body pose but not the initial velocity, the pelvis velocity takes about a half second to start tracking. Figure~\ref{fig:frames} shows a 3D rendering of several sample frames. See the supplementary materials for videos from more trials.

\begin{figure}[!htbp]
\centering
\includegraphics[width=1\linewidth]{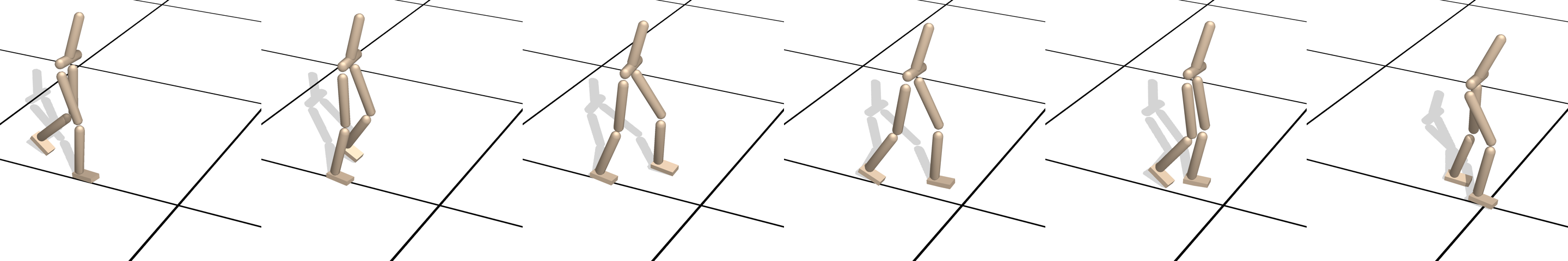}
\caption[]{Evenly spaced frames from one second of gait data from the same trial as Figure~\ref{fig:rollout}.}
\label{fig:frames}
\end{figure}

\subsection{Spatiotemporal metrics}

\begin{figure}[!htbp]
\centering
\includegraphics[width=1\linewidth]{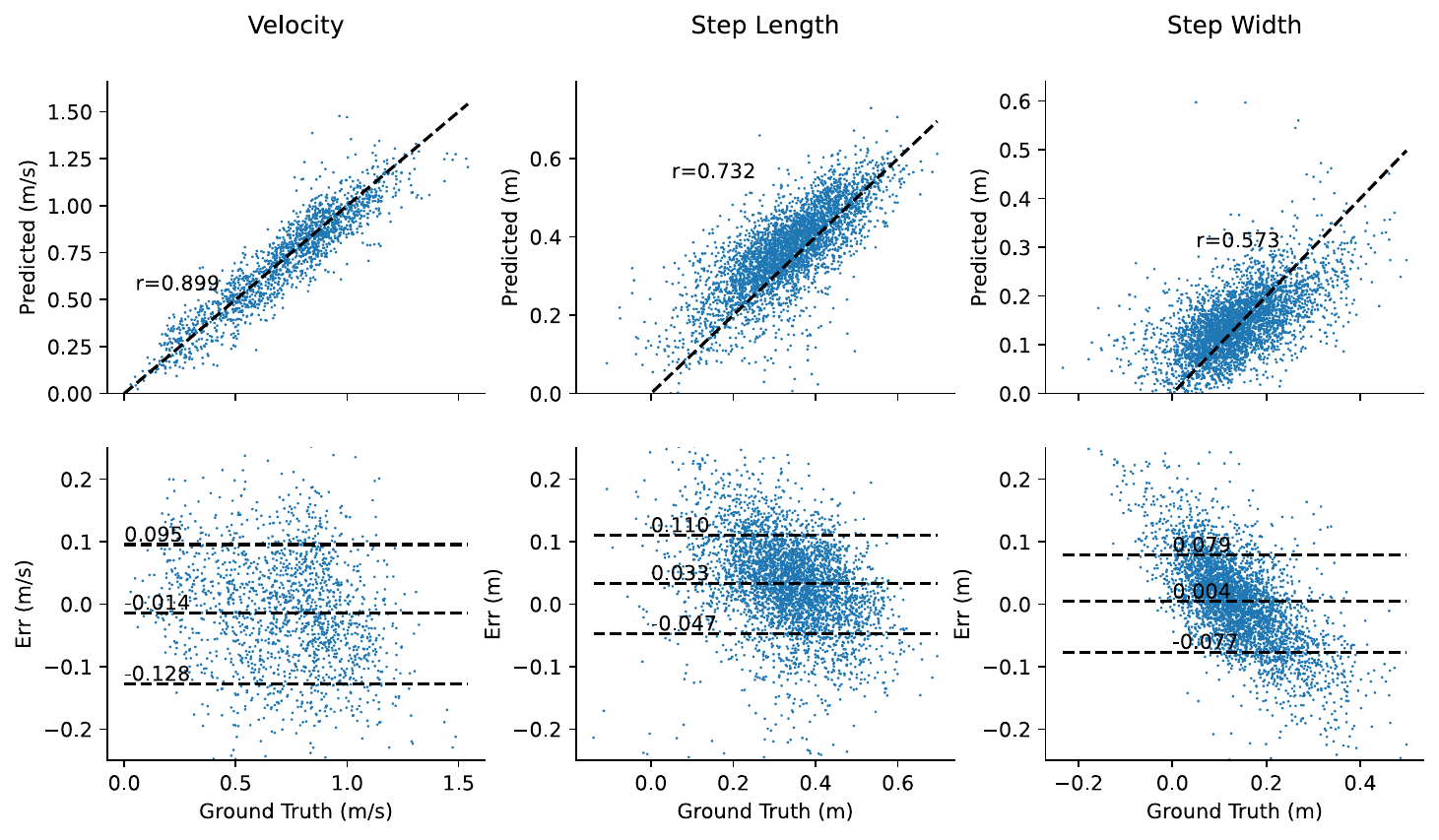}
\caption[]{Performance statistics estimating step length, step width, and velocity in the test data. Each metric is shown in a respective column. The top row show the correlations between the ground truth and predicted gait parameters. The bottom row shows the Bland-Altman plots, with the median difference and interquartile range indicated by horizontal bars.}
\label{fig:gait_params}
\end{figure}

Figure~\ref{fig:gait_params} plots the ground truth parameters for the step length, step width, and pelvis velocity averaged over a single gait cycle against the predicted values, with Bland-Altman plots for the errors with interquartile ranges. To quantify the performance, our primary metric of merit is the normalized interquartile range $\sigma_{NIRQ}=0.7413 \cdot \mathrm{IQR}$, which is comparable to the standard deviation of the errors, but is more robust to outliners. These occur in some trials where the model tracks poorly. We also report the RMSE for comparison to other works. These are reported in Table~\ref{table:metrics}. For our final model, $\sigma_{NIQR}$ was 0.094 m/s, 6.5cm, and 7.5 cm for velocity, step length, step width, respectively. These are estimates from a single gait cycle.

\subsection{Joint angles}

To quantify the performance tracking joint angles, we computed the root mean squared difference between the sagittal plane joint angles from the mocap data and the model tracking. This was 10.6° and 12.0° for the hip and knee respectively. Similarly to the gait transformer, we noted that these waveforms often showed a fixed offset relative to the ground truth data that would vary from trial to trial. Thus, we also computed the RMSE after removing the bias for each trial and the residual errors were 6.4° and 8.3°. The ankle joint was also included in our model and loss functions, but we found this performed rather poorly and did not regularly capture meaningful patterns. We suspect this was from a combination of an overly simplistic box model for the foot, the soft ground penetration allowed by the differential physics model, and the need to propel the model without proper tuning of masses and moments of inertia.

\subsection{Gait event timing}

\begin{figure}[!htbp]
\centering
\includegraphics[width=0.6\linewidth]{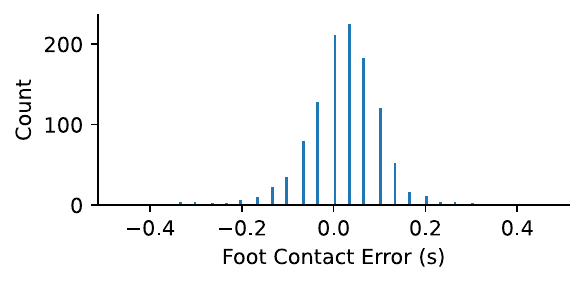}
\caption[]{Histogram of timing errors for foot contact events.}
\label{fig:contact_errors}
\end{figure}

We implemented a foot contact event detector based on the foot component penetrating the ground plane, which occurs with differentiable physics. We also applied slight thresholding of the contact force with a hysteresis loop, to prevent multiple detections for a given gait cycle. We then scored the accuracy of these foot contact events using Hungarian matching to the ground truth events and then measuring the timing error. A few events had detection errors greater than 500ms when the event was missed, so we discarded the 0.9\% of events outside that window. Figure~\ref{fig:contact_errors} shows the distribution of the timing errors. The RMSE for the residuals was 92 ms.

\subsection{Ablations and model comparisons}

After our model optimization using a combination of hyperparameter search through Weights and Biases and manual targeted search, we performed several model comparisons to highlight different design decisions. These results are all reported on the test set to compare against our final model. However, the model alterations are trained only on the 4200 trials that excluded the validation set from the full training set. Additionally, these results were only computed after we finalized the model hyperparameters, thus do not contaminate our cross-validation methodology.

Several optimal parameters jump out from these results. First, and as expected, including an initializer network to initialize the pose based on the first observation improves the results. We did not include this in most experiments as it requires differentiating through time and substantially slows down model training time (doubling to 9 hours). Removing the future timesteps from the observation also unsurprisingly reduced performance. Going from our default of 4 to 5 future timesteps also slightly improved performance, suggesting different architectures with greater future knowledge could yield greater gains. However, going to 10 future steps caused the performance to decrease. We also found using $r_{\mathtt {mse}}$ slightly outperformed $r_{\mathtt {exp}}$ by a small margin, although the later is standard in the literature. Using a PD controller also degraded performance. Our fairly large policy and value network sizes were identified through hyperparameter search, and we found reducing back to the smaller network that is the default setting of the Brax PPO implementation reduced performance.

To estimate how well this approach would work if not limited by any HPE inaccuracies, we trained a model where the mocap data was used in place of the keypoints in the observations. Controlling the model with the ground truth reduced the errors in all our metrics, but did not drive them to zero. We hypothesize some of this error arises from inconsistencies between the simple humanoid model we used and the more sophisticated biomechanical model used in the gait laboratory, which includes person-specific optimization of limb lengths. Our approach was only able to scale the model uniformly based on height.

\begin{table*}
\centering
\caption[]{Performance metrics for different models}
\scriptsize
\label{table:metrics}
\begin{tabular}{p{\dimexpr 0.083\linewidth-2\tabcolsep}p{\dimexpr 0.083\linewidth-2\tabcolsep}p{\dimexpr 0.083\linewidth-2\tabcolsep}p{\dimexpr 0.083\linewidth-2\tabcolsep}p{\dimexpr 0.083\linewidth-2\tabcolsep}p{\dimexpr 0.083\linewidth-2\tabcolsep}p{\dimexpr 0.083\linewidth-2\tabcolsep}p{\dimexpr 0.083\linewidth-2\tabcolsep}p{\dimexpr 0.083\linewidth-2\tabcolsep}p{\dimexpr 0.083\linewidth-2\tabcolsep}p{\dimexpr 0.083\linewidth-2\tabcolsep}p{\dimexpr 0.083\linewidth-2\tabcolsep}}
\toprule
Condition & Data & $\sigma^2_{IQR} Vel$ & RMSE Cycle Vel & $\sigma^2_{IQR}$ Step Len & RMSE Step Len & $\sigma^2_{IQR}$ Step Width & RMSE Step Width & RMSE Hip Angle & Hip Debiased & RMSE Knee Angle & Knee Debiased \\
\hline
final & train+val & 0.0942 & 0.1091 & 0.0649 & 0.087 & 0.0643 & 0.0747 & 10.5812 & 6.3972 & 12.0045 & 8.2596 \\
+init & train & 0.1021 & 0.1255 & 0.0645 & 0.0865 & 0.0594 & 0.0758 & 10.6561 & 6.434 & 12.1747 & 8.4267 \\
base & train & 0.1032 & 0.1114 & 0.0632 & 0.0854 & 0.0622 & 0.0761 & 10.7295 & 6.7478 & 12.1642 & 8.3374 \\
exp & train & 0.102 & 0.1206 & 0.0677 & 0.09 & 0.0585 & 0.0729 & 11.1396 & 6.843 & 12.0316 & 8.361 \\
-aug & train & 0.0979 & 0.1245 & 0.063 & 0.0875 & 0.0641 & 0.0738 & 10.7392 & 6.4789 & 12.0961 & 8.2182 \\
0 future & train & 0.1159 & 0.1429 & 0.0847 & 0.1005 & 0.0663 & 0.0793 & 11.6372 & 7.7355 & 13.676 & 10.2454 \\
5 future & train & 0.0979 & 0.1267 & 0.0614 & 0.0841 & 0.0581 & 0.0762 & 10.6977 & 6.5036 & 11.8867 & 8.1193 \\
10 future & train & 0.0984 & 0.119 & 0.0668 & 0.0949 & 0.0609 & 0.0864 & 10.3829 & 6.2094 & 11.4727 & 7.9165 \\
small & train & 0.1668 & 0.2383 & 0.1065 & 0.1325 & 0.1003 & 0.1143 & 14.8235 & 7.74 & 18.0834 & 10.4874 \\
mocap & train & 0.0414 & 0.113 & 0.041 & 0.0534 & 0.0393 & 0.0464 & 4.3263 & 3.8817 & 5.2813 & 4.7555 \\
\bottomrule
\end{tabular}
\end{table*}

\subsection{Comparison to Gait Transformer}

In comparison to our prior work, the Gait Transformer \citep{cotton_transforming_2022}, we found that incorporating physics into the inference process improved the accuracy of our estimates of velocity and step length.  With the gait transformer, the velocity estimates showed a correlation coefficient of 0.877 and an RMSE of 0.149 m/s and the step length showed a correlation of 0.704 and RMSE of 8.9 cm. In contrast, the correlations increased to 0.899 and 0.732 respectively and the RMSE decreased to 0.11 m/s and 8.7cm.  The heuristic we designed to detect foot contact events from the physics model had greater errors (mean absolute error 91 ms) than our gait transformer. (25 ms), although this was trained with more supervision on timing.  These results were measured on the same test dataset, although there are some slight differences in the trials included due to the filtering criteria and the need to discard the initial portion of the trial.

\section{Discussion}

We found that incorporating a simulated humanoid model into our gait analysis pipeline improves the accuracy of the estimated spatiotemporal gait parameters from monocular video. This includes estimating sagittal plane joint kinematics and step length from frontal plane videos, which is a challenging task. The correlations and RMSE for step length and velocity both outperformed our prior model-free algorithm \citep{cotton_transforming_2022} which used a transformer trained with supervised learning to map from keypoints to gait kinematics. This was also true for joint angles, although these still show errors that are too high. It remains an open question what the minimally clinically important difference is for many of these gait parameters in different patient populations.

However, there is still room for improvement. The errors that remain, even when the policy is driven by the motion capture data, show that our approach is unable to perfectly reproduce the observed movements. We suspect this arises because our humanoid model is overly simplistic, and because only scales the model proportionally with height rather than accounting for multiple anthropomorphic measurements. We intend to replicate this approach using more accurate biomechanical models in the future (e.g., \citep{caggiano_myosuite_2022, al-hafez_locomujoco_2023}). We anticipate including a more sophisticated anthropomorphic head that scales the model based on the keypoints or images will also increase the accuracy, versus our current approach scaling everything linearly.

Another potential benefit of such an anthropomorphic head would be to calibrate the mass and moments of inertia to match the ground truth torques and ground reaction forces. These are very clinically important but hard to estimate from kinematics alone and are available in our dataset. While our focus is on gait analysis, we anticipate training a large model over multiple diverse motion capture datasets and then fine-tuning it on clinical motion capture data will be critical for achieving this calibration while still generalizing. For example, a base model like PULSE \citep{luo_universal_2023} which can replicate all the movements in the AMASS dataset \citep{mahmood_amass_2019} without using residual force control would be a good starting point.  An additional benefit of PULSE is that it links imitation learning to generative modeling through a set of latent skills. This could allow generating representative gait patterns from gait representations discovered through self-supervised learning \citep{cotton_self-supervised_2023, winner_discovering_2022} and support data-driven attempts to learn the cost functions and constraints that give rise to different clinical patterns for walking. This could be further trained with the large datasets of movement from patients now becoming available with markerless motion capture \citep{cotton_markerless_2023}. Incorporating whole-body datasets would also allow tracking additional clinically important parameters such as whole-body angular momentum and extrapolated center of mass.

Another challenge was training a policy to precisely track a densely specified, high-dimensional goal, and we spent a lot of early effort on reward tuning. We are optimistic about recent approaches like Eureka \citep{ma_eureka_2023} that uses a language model to iteratively generate reward functions that outperform those written by domain experts. We also attempted to exploit the differentiable simulation of Brax using short horizon actor-critic \citep{xu_accelerated_2022}, which reports obtaining more precise policies than PPO. For our implementation at least, training was extremely slow and never achieved satisfactory performance. However, we do expect approaches that leverage the differentiable physics will be important for enabling more precise tracking after better aligning the humanoid model to biomechanical models.

A natural question is: why using reinforcement learning to introduce biomechanical and physical constraints into pose inference, in contrast to the traditional biomechanical pipeline that first estimates smoothed marker trajectories and then performs inverse kinematics on these trajectories? This was addressed in the context of autonomous drone racing \citep{song_reaching_2023}, which showed that decomposition into a trajectory planning and a trajectory following often lacks robustness to any model mismatch. In contrast, reinforcement learning more directly solves the specific problem: robust and precise trajectory following.

However, reinforcement learning does not elimitate the need for accurate pose estimates to drive the model, as indicated by the performance improvements when driving the model with ground truth data. We also found inconsistencies between the keypoints used and the body segments in the humanoid model a challenge. In contrast, many physics-based HPE tracking algorithms use models based on and driven by estimates of the parameters of the skinned multi-person linear (SMPL) model \citep{loper_smpl:_2015}. However, the kinematic model underlying SMPL is not designed for biomechanics. One promising solution to this is coupling physics-constrained inference with self-supervised learning of keypoint detectors \citep{gong_posetriplet_2022}. Furthermore, our approach to concatenating the keypoints for future frames and then directly feeding into the policy is likely a suboptimal architecture. We anticipate a temporal video-processing module that learns the optimal observation to drive the policy and is trained end-to-end with the reinforcement learning policy will outperform this design.

The performance we measured in this study may also not generalize to real-world use. Videos were all obtained from a fixed camera, and so when the camera is moving relative to the person we may see differing performance. These were also archival videos from clinical studies, and as such had been compressed to 640x480. Modern smartphones produce much higher-resolution videos. Finally, the gait laboratory dataset has an age distribution skewed towards children, and thus may not reflect the performance for adults or clinical populations not represented in the training data. Thus future work is required to validate this algorithm for clinical use, likely after implementing some of the improvements just described.

\section{Conclusion}

In conclusion, our study shows that incorporating physics-driven simulation of a humanoid model into the HPE inference process improves the performance of monocular video-based gait analysis. It outperforms our approach trained using model-free supervised learning and produces physically consistent inferences. Our work also highlights several opportunities for improvement ranging from more biomechanically precise models, inferring individualized anthropomorphic parameters, learning more precise policies for tracking fine details, and improving the video-based estimates for tracking.
\printglossaries




{\small
\printbibliography
}

\end{document}